\documentclass[letterpaper]{article} % DO NOT CHANGE THIS
\usepackage{aaai25}  % DO NOT CHANGE THIS
\usepackage{times}  % DO NOT CHANGE THIS
\usepackage{helvet}  % DO NOT CHANGE THIS
\usepackage{courier}  % DO NOT CHANGE THIS
\usepackage[hyphens]{url}  % DO NOT CHANGE THIS
\usepackage{graphicx} % DO NOT CHANGE THIS
\usepackage{natbib}  % DO NOT CHANGE THIS AND DO NOT ADD ANY OPTIONS TO IT
\usepackage{caption} % DO NOT CHANGE THIS AND DO NOT ADD ANY OPTIONS TO IT
\usepackage{algorithm}
\usepackage{algorithmic}

\usepackage{newfloat}
\usepackage{listings}
\DeclareCaptionStyle{ruled}{labelfont=normalfont,labelsep=colon,strut=off} % DO NOT CHANGE THIS
\floatstyle{ruled}
\newfloat{listing}{tb}{lst}{}
\floatname{listing}{Listing}
\usepackage{amsfonts}
\usepackage{amsmath}
\usepackage{multirow}

\title{Learning Multi-Modal Whole-Body Control \\ for Real-World Humanoid Robots}

\author{
  Pranay Dugar, Aayam Shrestha, Fangzhou Yu, Bart van Marum, Alan Fern
}

\affiliations{
  \{dugarp, shrestaa, yufangzh, vanmarub, afern\}@oregonstate.edu \\
  Collaborative Robotics and Intelligent Systems Institute (CoRIS) \\
  Oregon State University \\
  Corvallis, Oregon 97331, USA
}

\copyrighttext{Accepted at AAAI Fall Symposium. \\This work is supported by NSF Award 2321851, DARPA contract HR0011-24-9-0423, and the NVIDIA Academic Grant Program.}

\begin{document}

\maketitle

\begin{abstract}
A major challenge in humanoid robotics is designing a unified interface for commanding diverse whole-body behaviors, from precise footstep sequences to partial-body mimicry and joystick teleoperation. We introduce the Masked Humanoid Controller (MHC), a learned whole-body controller that exposes a simple yet expressive interface: the specification of masked target trajectories over selected subsets of the robot's state variables. This unified abstraction allows high-level systems to issue commands in a flexible format that accommodates multi-modal inputs such as optimized trajectories, motion capture clips, re-targeted video, and real-time joystick signals. The MHC is trained in simulation using a curriculum that spans this full range of modalities, enabling robust execution of partially specified behaviors while maintaining balance and disturbance rejection. We demonstrate the MHC both in simulation and on the real-world Digit V3 humanoid, showing that a single learned controller is capable of executing such diverse whole-body commands in the real world through a common representational interface.

\end{abstract}

\begin{links}
    \link{Project Webpage}{https://masked-humanoid.github.io/mhc/}
\end{links}

\section{Introduction}
Humanoid robots hold immense potential as highly capable and adaptive platforms for tackling complex real-world tasks, thanks to their dexterous multi-purpose body structure that mirrors our own. However, development of versatile and robust whole-body controllers for bipedal humanoids remains a critical challenge in robotics. Traditional approaches involve meticulous manual engineering of separate controllers for different skills such as standing \cite{crowley2023optimizing}, walking \cite{castillo2021robust}, manipulation \cite{dao2023sim} and mimicry \cite{he2024learning}, resulting in specialized controllers with limited versatility and adaptability. While effective in isolation, these controllers lack versatility, are difficult to extend, and rarely support seamless reuse across tasks.

A truly versatile whole-body humanoid controller should exhibit several key properties. First, it should be able to robustly stand, walk, and mimic whole and partial-body motions. Second, it should comprehend and execute these capabilities through \emph{multiple input modalities}, including combinations of high-level velocity commands, end-effector targets, video demonstrations, or motion capture data. Third, the controller should be \emph{robust} to dynamic command updates, noisy or inexact inputs, inaccurate simulation parameters, and external disturbances. Finally, the controller should allow for straightforward extension of its repertoire as new motion examples become available, with no or minimal retraining.

A key challenge is the lack of a unified representation for specifying robot behaviors. Tasks may be defined through velocity commands, end-effector targets, human demonstrations, or partial-body trajectories, yet most systems rely on distinct processing pipelines for each modality. This fragmentation slows application development, restricts interoperability, and complicates the transfer of policies from simulation to real hardware.

To address these challenges, we train the Masked Humanoid Controller (MHC), a whole-body controller that provides a unified representational interface for multi-modal humanoid control. MHC represents target behaviors as future pose trajectories with optional masking, allowing the same framework to accommodate velocity-based walking, arm-only imitation, whole body imitation or combined loco-manipulation tasks.  We train the MHC via reinforcement learning using a carefully designed curriculum and an expansive library of behaviors spanning optimized reference trajectories, re-targeted video clips, and human motion capture data. The tailored curriculum over motion commands and disturbances gradually introduces capabilities with the aim of achieving the above desired robustness and versatility properties. 
\begin{figure*}[t]
  \centering
  \includegraphics[width=\textwidth]{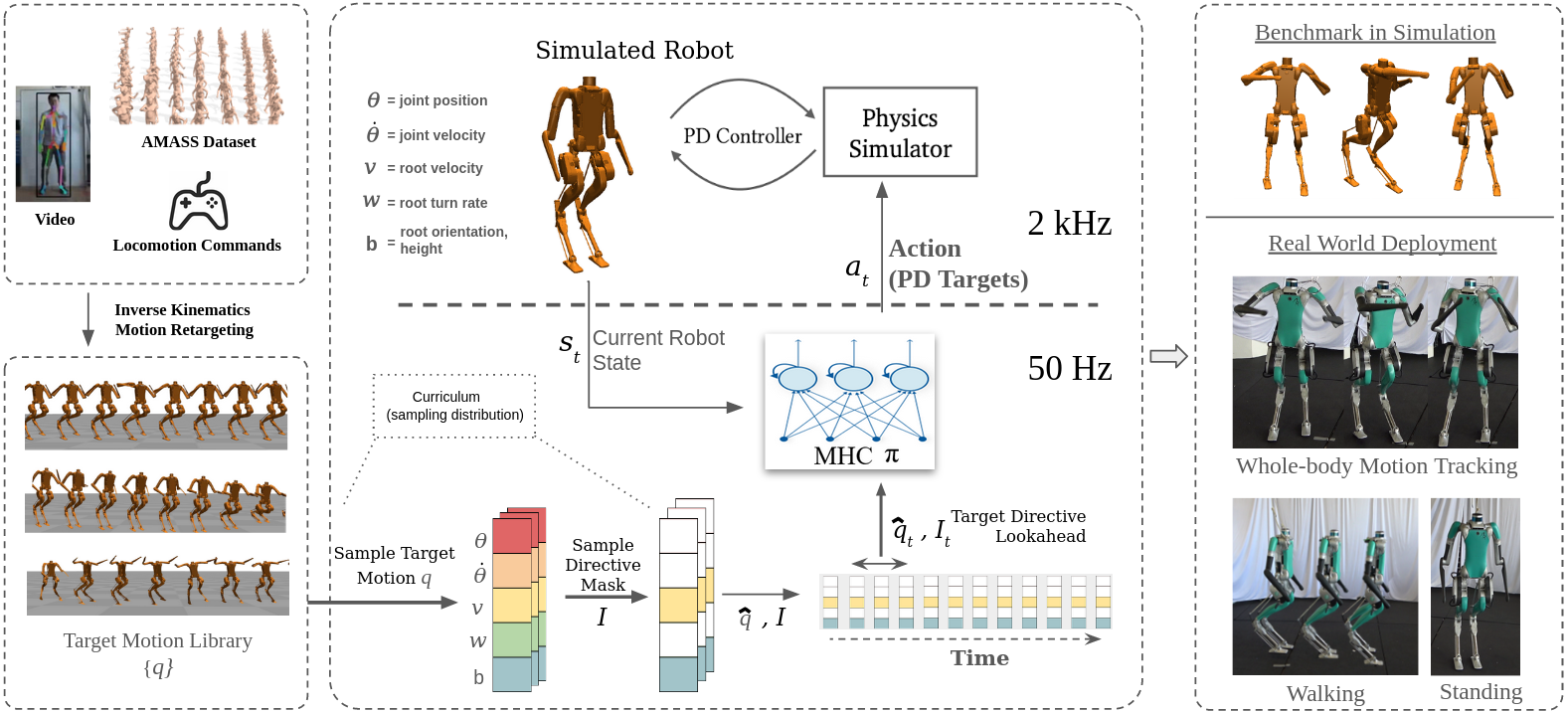}
  \caption{
The Masked Humanoid Controller (MHC) is learned from a dataset of re-targeted human motions paired with torso locomotion commands, including standing. During training and testing, masking can be applied to target motion trajectories to yield masked motion directives that are given to the MHC. The MHC then produces PD setpoints for the whole body in order to track the current motion directive. Training includes domain randomization and force perturbations to facilitate robustness and transfer from simulation to the real robot.}
  \label{figure:main}
\end{figure*}

We validate the MHC through simulated benchmarking and showcase qualitative and quantitative results for sim-to-real transfer using the Digit V3 robot platform. The simulation experiments demonstrate the importance of the curriculum, architecture choices and generalization to new motions. Our real-world experiments demonstrate the robustness of the MHC across a variety of target behaviors such as walking, boxing, and box loco-manipulation, while also highlighting current shortcomings in sim-to-real transfer. 

\section{Related Work}

\textbf{Motion Tracking for Simulated Humanoids.} Motion capture data have been used extensively for generating motions of simulated characters. Most closely related to our work is the application of reinforcement learning with tracking rewards over fully specified target motions \cite{Won2022PhysicsbasedCC, luo_perpetual_2023, Merel2018NeuralPM, Dou2023CASELC, wagener_mocapact_2023, Peng2022ASE}. However, tracking rewards alone do not always yield smooth transitions between skills and failure recovery. Recent works augment training with adversarial losses to encourage natural motions during transitions \cite{Peng2022ASE, tessler2023calm} or define an explicit fail state recovery policy \cite{luo_perpetual_2023}. While some approaches offer more intuitive control for under-specified tasks, they often focus on predefined types of input trajectory sparsity, and adapting to new sparsity specifications, such as VR, require retraining \cite{Cern2023ANM, Du2023AvatarsGL}. Notably, existing models designed for physically realistic animation use the SMPL skeleton in IsaacSIM, such as PHC \cite{luo_perpetual_2023}, and ASE \cite{Peng2022ASE}, and require substantial engineering effort and fine-tuning to adapt to a real humanoid in the MuJoCo setup with dynamics randomization. Moreover, discrepancies between simulated and physical environments—such as differences in environmental physics and unrealistic joint definitions in simulated characters—can result in behaviors that are not physically plausible or efficient for actual robots. Work such as Versatile Motion Priors (VMP) \cite{serifi2024vmp} learns latent motion embeddings from unstructured human motion and trains dynamics-aware policies that can generalize to diverse behaviors on real humanoids, showing promise for bridging the simulation-to-reality gap. Recent work on the Masked Humanoid Controller (MHC) \cite{shrestha2024mhc} addresses these challenges by learning a unified low-level policy from sparsely masked motion data, enabling robust tracking, smooth transitions, and generalization across diverse, under-specified input modalities.  Our work similarly aims to incorporate a wide variety of sparsity types in a single controller while focusing more on addressing the simulation to reality gap.

\textbf{Robust Locomotion Control.} Standing, walking, and transitioning from one to the other present distinct challenges in stability and adaptability, driving recent advances in adaptive control strategies that enable robots to navigate complex terrains and perform dynamic maneuvers. Significant progress has been made in the locomotion of quadrupeds, demonstrating enhanced stability \cite{lee2020learning, da2021learning} and agility \cite{jenelten2019dynamic} in various environments. \cite{fu2023deep} train a unified whole-body manipulation and locomotion controller that attaches an arm on top of a quadruped to perform mobile manipulation tasks. Notable achievements for bipedal locomotion with the Cassie platform (no upper torso or arms) include blind locomotion across multiple gaits and behaviors \cite{siekmann2021sim, li2021reinforcement}, locomotion under varying loads \cite{dao2022sim}, and visually-guided locomotion over irregular terrain using deep learning to process visual input and adapt locomotion strategies accordingly \cite{duan2023learning}. More recently a learned controller for blind locomotion on the full-body humanoid Digit was demonstrated across varying terrain and disturbances \cite{radosavovic2023learning}. These advancements underscore the importance of robust data and simulation technologies in developing and refining control strategies. However, managing transitions between different modes of locomotion, such as standing and walking, remained a significant challenge. Van Marum et al. \cite{van2024revisiting} addressed this complexity with the development of the SaW controller for Digit, which integrates stability and adaptive walking strategies into a single operational framework that maintains balance against natural disturbances. However, standing and walking alone do not constitute whole-body control. 

\textbf{Mimicry for Humanoids.}
Efforts to closely replicate human motion and interaction within robotic frameworks have been significantly enhanced by both simulation techniques and biological inspirations. Early works by Dariush et al. \cite{dariush2008whole} and recent advancements by Li et al. \cite{li2024reinforcementosu} have advanced the field of humanoid motion tracking for locomotion, emphasizing the utilization of optimized trajectories and multi-stage training setups to facilitate transfer from simulation to the real world. In concurrent work, He et al. \cite{he2024learning} have developed a controller trained on the AMASS dataset \cite{Mahmood2019AMASSAO} to learn through mimicry, another key modality of control. While this approach effectively performs whole-body control, it only covers the range of locomotion capabilities present in the dataset. As a result, the system's performance is constrained by the locomotion patterns and upper body limitations observed in the mimicry data. They do not demonstrate robust joystick-style locomotion capabilities that are required for loco-manipulation tasks. 

\textbf{Upper-Body Motion Tracking.} Recent research by Cheng et al. \cite{cheng2024expressive} covers the modality of partial-body control by performing upper-body mimicry from target motions and joystick-based lower-body control. It cannot however handle the modalities of standing or whole-body mimicry. It performs a continuous step-in-place motion when commanded for $0$ velocity and lacks height control beyond the region of its nominal height. These are properties that are crucial for whole-body mimicry when considering lower body. It also necessitates the need for commanding a motion for upper-body even while performing simple locomotion tasks. An unconstrained locomotion controller would be able to learn optimal arm swing motions that are speed-dependent and perform better under disturbances as the arms would have more mobility when left unconstrained. 

The vast majority of the above approaches focus on a single input modality, such as walking, standing, partial-body mimicry, or whole-body mimicry. In order to jointly handle all these input modalities, these approaches would need to either switch between controllers or have an interpreter that can generate a whole-body motion for different speeds and optimal arm positions for the commanded speed on the fly. This can result in significant added complexity. Concurrent work such as HOVER \cite{he2024hover} demonstrates a unified policy architecture capable of handling multiple control modes without explicit switching by distilling diverse skills into a single representation space. Our MHC follows a similar philosophy, integrating locomotion and comprehensive whole-body motion tracking in humanoids into a single unified framework. It enables natural and adaptable motion generation that a user can control based on the scenario at hand with no overhead of policy switching or complex interpreters.

\section{Problem Formulation}

Given a dataset of humanoid motions from various behaviors, our objective is to learn a controller which can match target motion directives that are representative of the data distribution. We are particularly interested in supporting partially-specified motion directives e.g. only specifying the upper-body joint trajectories, or just the torso velocity, or both. It is expected that the controller ``fills in the blanks" for joints that are not specified by a directive, e.g. details of the lower-body joints.
Such a controller can support directives derived from various input modalities. For example, fully-specified directives specifying all joints can be derived from MoCap data, while joystick commands regarding the root velocity and arm movements correspond to partial directives. 

More formally, a \emph{motion} is a sequence of poses $q_{1:H}$ for a humanoid with $J$ joints over $H$ time steps. Each pose is represented by a tuple $q_i = (\theta_i, {\dot{\theta}}_i, v_i, w_i, b_i)$, where $\theta_i \in \mathbb{R}^J$ denotes the joint angles, ${\dot{\theta}}_i \in \mathbb{R}^J$ denotes the joint angular velocities, $v_i \in \mathbb{R}^2$ denotes the root planar linear velocities, $w_i \in \mathbb{R}$ denotes the turn rate for the humanoid, and $b_i \in \mathbb{R}^3$ denotes the Euler base orientation in the x-axis, y-axis, and the height of the base from the ground plane. Motion directives are used to specify constraints on a desired motion to be generated. Specifically, a \emph{motion directive} $d$ is defined as a masked motion sequence represented by $d = (\hat{q}_{1:H}, I_{1:H})$, where $\hat{q}_{1:H}$ is a masked motion and $I_{1:H}$ is a sequence of binary masks such that $I_i$ indicates which dimensions of pose $q_i$ are selected as motion constraints. We follow the convention of setting the masked dimensions of $\hat{q}$ to zero. Note that while the definition of a directive allows for arbitrary sets of state variables to be masked, in practice, we focus training and evaluation on masking patterns relevant to the multiple input modalities the controller needs to support as described in Section \ref{sec:training}.

The Masked Humanoid Controller (MHC) is a controller $\pi$ that at each time step $t$ takes an input containing the current \emph{humanoid proprioceptive state} $s_t = (r\theta_t, \dot{r\theta}_t, \omega_t)$ and a target motion directive $d = (\hat{q}_{1:H}, I_{1:H})$, where $r\theta_t \in \mathbb{R}^J$ and $\dot{r\theta}_t \in \mathbb{R}^J$ are the joint positions and velocities, and $\omega_t$ is the torso orientation in quaternion form. Note that $s_t$ is composed of the actual robot states while $d$ is composed of the motion being commanded. The output of the MHC is an action $a_t = \pi(s_t,d)$, which in our work corresponds to PD setpoints for all robot motors. The objective of the MHC is to select actions which aligns the future humanoid motion with the commanded directive, while maintaining stability and robustness to disturbances. 

This formulation allows a single controller to generalize across diverse input modalities, supporting both robotic research workflows and artistic motion design. For example, specifying only torso velocity and arm trajectories enables the controller to autonomously generate stable leg motions, while fully-specified directives from motion capture ensure high-fidelity imitation. This unified representation removes the need for multiple specialized controllers or interpreters.

\section{Masked Humanoid Controller}
\label{sec:training}

Due to the lack of low-level supervisory information, we train the MHC via reinforcement learning (RL) in simulation before transferring it to the real world. The MHC is trained using the PPO algorithm in a MuJoCo physics engine simulation with a Digit V3 humanoid model. Below we detail our training approach which involves several key components: motion dataset generation, MHC network architecture, a curriculum of training episodes, domain randomization for sim-to-real transfer, and our mask aligned reward function. 

\subsection{Motion Data Generation} 
\label{subs:motiongen}
We create a diverse set of reference motions for training the MHC using a combined dataset consisting of human motion capture (MoCap) datasets (AMASS \cite{Mahmood2019AMASSAO}, Reallusion~\cite{Reallusion2022}), video demonstrations, and optimized trajectories designed using spring mass model of Digit. To help bridge the human to Digit embodiment gap, we employ an Inverse Kinematics (IK) based retargeting procedure to map the dataset trajectories to Digit's kinematic model. 

For each frame in the dataset, we solve an IK problem for the generalized position vector of digit $q=(\theta, b)$ formulated as a nonlinear program (NLP). We use the IK module in Drake \cite{drake} to set up the costs and constraints associated with the NLP, and SNOPT \cite{gill2005snopt} as the underlying solver. Critical kinematic feasibility requirements such as locking the stance foot to the ground and respecting the closed kinematic chain topology of Digit's legs are expressed as constraints in task space. Upper body motion targets that serve more of a stylistic purpose such as torso pose, and hand positions are expressed using costs to aid in convergence. Any frames that the IK optimization did not generate a feasible solution for are simply dropped. Linear interpolation is applied to the sequence of states $q_i$ and their timestamps $t_i$ to produce the final time-parameterized trajectory $q_{1:H}$ used for training. 

We note that there are some discrepancies in retargeted motions when MoCap trajectories are not kinematically feasible for the Digit robot. This is due to Digit's one-dimensional rotational joints (limited to $<360^\circ$ rotation) vs MoCap's 3D spherical joint models. The retargeting process uses inverse kinematics (IK) with cost-based optimization to find the best approximate solution within these constraints.

\subsection{Network Architecture} 
The MHC model is a Long Short-Term Memory (LSTM)~\cite{lstm} recurrent neural network which takes as input the current humanoid state $s_t$ and the next step of the current directive $d_i=(\hat{q}_i, I_i)$. The MHC first processes the directive with a single-layer feed-forward encoder to produce a 160 dimensional directive embedding. This is concatenated with $s_t$ and fed into an LSTM block configured with two 64-unit recurrent layers to capture temporal information. A final linear decoding layer outputs an offset for each actuated joint. The MHC action $a_t$ is computed by adding this offset to the actuated joint values in $\hat{q}_i$, which yields the next PD setpoints. Note that for the masked joints in $d_i$, the corresponding values in $\hat{q}_i$ are zero and the offset corresponds to the actual PD setpoint. In our work, we use the Digit V3 humanoid, which has 20 actuated joints. The MHC is run at 50Hz to compute PD setpoints which are sent to a PD controller running at 2kHz. 

\subsection{Training Episode Generation} 
Each episode is initialized with the robot in a standing position. Next a random command window length $w$ is generated and a random directive $d$ is drawn from a distribution defined by the curriculum stage (see below). The MHC then cycles through $d$ until reaching $w$ steps. This sampling of a $w$ and $d$ continues until the robot falls or reaches the maximum episode length $e$, which depends on the curriculum stage. By switching between multiple random directives during an episode, the MHC can learn to smoothly transition between different types of motions. In addition, to encourage robustness, during the execution of each episode we apply perturbations on the torso according to a distribution that depends on the curriculum stage. 

\subsection{Curriculum Stages} We employ a three-stage curriculum that progressively focuses training on locomotion, robustness to perturbations, and whole-body motion tracking. This enables gradual acquisition of complex skills while ensuring stable learning.

\textbf{Stage 1 - Locomotion:} The objective of this stage is to focus learning on developing basic balance and locomotion control, which is a core capability required for more complex behaviors. For this purpose, the episodes involve only randomized \emph{locomotion directives}, which are partial directives that mask all variables except for those specifying root motion $(v_i, w_i, b_i)$. Instantaneous torso perturbations (80-800N) are also introduced, each affecting a single policy step with a $1\%$ probability. Training lasts for 300 policy steps (6 seconds) with a command window $w \in [40, 100]$.

\textbf{Stage 2 - Stability:} Building upon Stage 1, we focus on enhancing the controller's stability during locomotion. In this stage continuous torso perturbations (20-150N) are applied over windows of 20-50 policy steps. In order to facilitate recovery, we increase episode length to 800 policy steps (16 seconds) and the command window length to $w \in [100, 400]$.

\textbf{Stage 3 - Whole-Body Directives:} In this stage, we train the MHC to follow both fully and partially specified directives. Fully-specified directives provide the complete motion reference without any masking, requiring the MHC to perform whole-body imitation. Partially specified directives may take three forms, which correspond to key input modalities: 1) providing only locomotion commands without any upper or lower-body motion specification, corresponding to a joystick input modality, 2) specifying only the upper-body motion while standing in place, upper body imitation from VR or video modalities, or 3) specifying upper-body target motions with locomotion commands, corresponding to a richer VR controller with a locomotion interface. The MHC learns to generate coherent whole-body motions while satisfying the constraints specified by the unmasked components of the directive. Throughout this stage, we maintain the episode length at 800 policy steps and the command window length $w$ within the range [100, 400].

\subsection{Domain Randomization} 
To facilitate transfer from simulation to the real robot, we employ dynamics randomization throughout all stages of the curriculum. This involves randomizing various physical parameters of the simulated environment, such as joint damping, link masses, center of mass positions, encoder noise, ground friction. By exposing the MHC to a wide range of dynamics during training, we aim to learn a policy that can generalize to the dynamics of the real robot without requiring precise system identification or extensive real-world fine-tuning. The randomization ranges are carefully chosen to strike a balance between creating sufficiently diverse training scenarios and maintaining realistic behavior of the simulated robot. We use the same randomization ranges as prior work on Digit locomotion training \cite{van2024revisiting}.

\subsection{Reward Design}
\label{sec:reward}
Our reward function must encourage stability and robustness during locomotion and standing as well as tracking of unmasked joints in the current directive. For locomotion and standing stability we build upon the reward function used to train a recent state-of-the-art humanoid standing and walking controller \cite{van2024revisiting}. That reward function includes three components: 1) \textit{Task Reward} $r_{task}$ to encourage command following for locomotion directives, 2) \textit{Style Reward} $r_{style}$ to encourage natural body postures for standing and walking, 3) \textit{Regularization Reward} $r_{reg}$ to encourage smooth joint motion and low torques. We refer the reader to the original paper for details of these components. We build on these components and add a \textit{Tracking Reward} $r_{track}$.
$$r_{track} = \exp\left(-1.5 \cdot || (\theta - \hat{\theta}_i) \cdot I_i||\right),$$    
where $\theta$ is the current joint values and $\hat{\theta}_i$ is the target joint for that step along with the corresponding directive mask $I_i$.

The overall reward function depends on whether the directive masks the lower-body joints or not. When the lower-body is masked, e.g. for pure locomotion commands or locomotion combined with upper body tracking, the reward function is $r = r_{task} + r_{style} + r_{reg} + r_{track}$. However, when the lower-body joints are not masked in the directive, the reward function removes the style reward and becomes $r = r_{task} + r_{reg} + r_{track}$. This is done because the motion preferences of the style reward may conflict with the specific lower-body target directives that should be tracked. 

\section{Experimental Results}
\label{sec:result}

We present our evaluation protocol and simulation experiments to assess the MHC's performance. We  compare the MHC with baseline approaches, analyze the impact of varying training sets, and learning curriculum. Additionally, we present qualitative and quantitative results of sim-to-real transfer to the Digit V3 robot.

\renewcommand{\arraystretch}{1.1}
\begin{table}[t]

\centering
\begin{tabular}{|l|l|}
\hline
\textbf{Component} &  \textbf{Range} \\
\hline
x velocity ($v_{ix}$) & [-0.5, 2.0]m/s \\
y velocity ($v_{iy}$) & [-0.5, 0.5]m/s \\
turn rate ($w_i$) & [-0.5, 0.5]radians/s \\
torso x orientation ($b_{ix}$) & [-0.314, 0.314]radians \\
torso y orientation ($b_{iy}$) & [-0.314, 0.314]radians \\
torso height ($b_{ih}$) & [0.5, 1.0]m \\
Standing mask bit ($I_{is}$) & \{0, 1\}\\
\hline
\end{tabular}
\caption{Predefined ranges for components of locomotion directives. Velocities ($v_{ix}, v_{iy}$) and turn rate ($w_i$) are set to zero when the standing mask bit ($I_{is}$) is set to $1$, indicating a standing directive.}
\label{table:loc_ranges}
\end{table}

\subsection{Motion Directive Dataset}
We used IK retargetting to generate 75 diverse kinematic motion trajectories. We selected this set of motions based on considering diversity, fit to the robot morphology, and avoiding overly aggressive motions such as jumping, kicking, and high velocity swinging of limbs. In order to study the generalization capabilities of our approach, we test MHC models on motions outside of their training sets, dividing the data into three sets, each containing a mix of motions from the original motion sources. 

\begin{itemize}
\item \textit{setA} [20 motions]: Amass Boxing (5), Amass misc (9), Reallusion (2), optimized (2), video (2) 
\item \textit{setB} [20 motions]: Amass Boxing (6), Amass misc (6), Reallusion (5), optimized (3) 
\item \textit{setC} [35 motions]: Amass Boxing (19), Amass misc (12), Reallusion (1), optimized (3) 
\end{itemize}

While each set contains simple upper-body movements (such as waving), the general trend is an increasing level of difficulty as judged by the authors. We note that \textit{setC} has a number of the most difficult motions, involving highly dynamic motion, such as tennis smashes and difficult boxing moves requiring precise and dynamic footwork combined with upper body motions. The feasibility of these more difficult motions is not obvious for our realistic Digit model.

\subsection{Experimental Setup and Metrics}

We evaluate the MHC on different input modalities including partial/whole-body mimicry, walking, and standing with perturbations. We use the full re-targeted dataset for whole-body mimicry and mask out the lower body targets for partial-body mimicry. Each episode for full/partial mimicry experiments command the complete motion directive once. 

For pure walking experiments the directives are sampled from a predefined range as specified in Table \ref{table:loc_ranges}. For each walking experiment episodes last for 10s and all non-locomotion directive components are masked out. Finally, for standing perturbation testing we follow prior work \cite{van2024revisiting} and provide an input directive of standing and apply varying forces at the base of the torso for a window of length of 25 steps to measure the root position drift. Below we describe the metrics used throughout our evaluations.

\begin{itemize}
    \item \emph{Mean End-Effector Positional Error} ($\mathbb{E}_{\text{MEPE}}$): Mean positional error for end effectors (hands, elbows, knees, feet) is calculated as the L2 norm of the current Euler position relative to the torso, compared to values from the directive. For partial directives, the error is computed only for unmasked joints. 
    %This is done for both partial and fully-specified directives before episode termination.
    \item \emph{Root Drift} (Root$_\Delta$): The mean of drift in root position during an episode from its commanded position when following partially or fully-specified directives. Computed as an L2 norm of the current root position against expected root position in the previous directive step. 
    \item \emph{Failure Rate} (Fail \%): Percentage of episodes that result in failure (i.e. falling). 
    
\end{itemize}

\subsection{Comparison to Baselines} 

Since no prior approaches handle masked directive inputs like the MHC, we compare our model with different baselines for locomotion and mimicry. For locomotion, we compare against a state-of-the-art standing and walking controller\cite{van2024revisiting}. We also develop two natural baselines to compare against for locomotion + upper-body mimicry: 1)\textit{OffsetBlind}. This model focuses on solely tracking locomotion components of input directives and does not receive upper-body IK offsets as model input. Rather it feeds upper-body offsets from the directive straight to the PD controllers without considering them in its locomotion planning.
2) \textit{OffsetAware}. This model is the same as OffsetBlind, except that it receives upper-body offsets in its input. This allows it to adjust its locomotion control based on the anticipated  upper body motions. 

\begin{table}[t]
\centering
\setlength{\tabcolsep}{2mm} % Slightly compress columns if needed
\begin{tabular}{|l||c|c||c|c|}
\hline
\multirow{2}{*}{\textbf{Model}} & \multicolumn{2}{c||}{\textbf{Walk Root$_\Delta$ (m)}} & \multicolumn{2}{c|}{\textbf{Stand Root$_\Delta$ (m)}} \\ \cline{2-5}
 & 1x m/s & 0.5y m/s & 50N & 100N \\ \hline \hline
\textbf{MHC}   & 0.413 & 0.343 & 0.256 & 0.311 \\ \hline
\textbf{SaW(SOTA)}  & 0.803 & 0.388 & 0.109 & 0.452 \\ \hline
\end{tabular}
\caption{Locomotion performance comparison of MHC models with the SOTA Stand-and-Walk (SaW) controller. Root drift ($\mathrm{Root_\Delta}$) is measured for 10\,s walking episodes and standing under applied torso forces.}
\label{table:loc_noisy_loco_res}
\end{table}

\renewcommand{\arraystretch}{1.3}
\begin{table}[t]
\centering
\begin{tabular}{|l|r|r|r|r|r|r|}
\hline
             \textbf{Model} &   \textbf{Fail \%} & \textbf{$\mathbb{E}_{\text{MEPE}}$(in m)} & \textbf{Root$_\Delta$(in m)} \\
\hline
\textbf{OffsetBlind} & 49.213	& 0.156	& 0.168 \\
   \textbf{OffsetAware} & 4.800	& 0.227	& 0.203 \\
               \textbf{MHC} & 0.450 &	0.098 &	0.141 \\
\hline
\textbf{Real World} & 5.0 &	0.117 & - \\
\hline
\end{tabular}
\caption{Comparison of baseline models on various metrics, including failure rate, mean per-joint position error ($\mathbb{E}_{\text{MEPE}}$), and root mean squared error (Root$\Delta$). The MHC model achieves the lowest failure rate and error metrics compared to the OffsetAware and OffsetBlind models. Root$\Delta$ for Real not computed due to missing global position data per frame.}
\label{table:input_result}
\end{table}

For this comparison experiment we train the MHC and all baselines on datset \textit{SetA}. The results of our standing and walking evaluation in comparsions to the SOTA \cite{van2024revisiting} are provided in Table \ref{table:loc_noisy_loco_res}. We give results for the root drift when given walking commands (+1m/s x and +0.5m/s y) and for standing under different force perturbations (50N and 100N). The MHC demonstrates competitive performance in both walking and standing tasks, highlighting that MHC is on par with SOTA walking and standing capabilities.
 
Table \ref{table:input_result} gives results comparing the MHC to the mimicry baselines. The MHC outperforms the baselines across all metrics. Most significantly it achieves a failure rate of less than 1\% which is significantly lower than the baselines. Interestingly, we see that OffsetAware achieves a significantly lower failure rate than OffsetBlind, showing the utility of explicitly training the the controller to prepare for upcoming target motions provided in its input. 

Tables \ref{table:loc_noisy_loco_res} and \ref{table:input_result} evaluate complementary aspects of MHC performance. Table \ref{table:loc_noisy_loco_res} isolates locomotion robustness, comparing MHC against the state-of-the-art SaW controller under force perturbations. Table \ref{table:input_result} focuses on whole-body imitation with locomotion integration, comparing MHC to mimicry baselines.

\subsection{Evaluation of Generalization Performance}

We train a different MHC on three varying sets of motions to investigate its generalization and influence of different training sets. Here MHC$_{\text{A}}$ is trained on \textit{setA}, MHC$_{\text{AB}}$ on \textit{setA} \& \textit{setB}, and MHC$_{\text{ABC}}$ on all three sets. We then evaluate these models on each dataset and report the performances in Table \ref{table:gen_res}. Our results demonstrate two key findings:

\textbf{Generalizability:} All three models consistently perform well on the datasets they were trained on, regardless of the dataset size. MHC$_{\text{A}}$ excels on \textit{setA}, MHC$_{\text{AB}}$ shows strong performance on both \textit{setA} and \textit{setB}, and MHC$_{\text{ABC}}$ performs well across all three sets. We also see that MHC$_{\text{A}}$ and MHC$_{\text{AB}}$ yield strong performance on the datasets they were not trained on. This indicates the generalization capability of the MHC to motions outside of its direct training experience. 

\textbf{Improved Performance with Larger Datasets:} We observe a trend of increasing performance as the training dataset size grows. Notably, models trained on larger datasets often outperform those trained on smaller subsets, even when evaluated on the smaller sets. For instance, MHC$_{\text{AB}}$ shows better performance on \textit{setA} compared to MHC$_{\text{A}}$, and MHC$_{\text{ABC}}$ achieves the best results on \textit{setB} and \textit{setC}. This is good evidence that with additional compute the MHC has the potential to continue improving as additional motions sources are incorporated into training.

\renewcommand{\arraystretch}{1.3}
\begin{table}[t]
\centering
\begin{tabular}{|l||c|c|c||c|c|c|}
\hline
\multirow{2}{*}{\textbf{Metric}} & \multicolumn{3}{c||}{\textbf{$\mathbb{E}_{\text{MEPE}}$(in m)}} & \multicolumn{3}{c|}{\textbf{Root$_\Delta$(in m)}} \\ \cline{2-7}
 & setA & setB & setC & setA & setB & setC \\ \hline\hline
\textbf{MHC$_{\text{A}}$}   & 0.10 & 0.57 & 0.82 & 0.14 & 0.55 & 0.44 \\ \hline
\textbf{MHC$_{\text{AB}}$}  & \textbf{0.09} & 0.55 & 0.78 & \textbf{0.09} & 0.44 & 0.41 \\ \hline
\textbf{MHC$_{\text{ABC}}$} & 0.10 & \textbf{0.49} & \textbf{0.75} & 0.10 & \textbf{0.37} & \textbf{0.40} \\ \hline
\end{tabular}
\caption{Performance Comparison of MHC models on partial and whole-body directives computed on different datasets (\textit{setA}, \textit{setB}, \textit{setC})}
\label{table:gen_res}
\end{table}

\subsection{Curriculum Results}

We also train and evaluate \textit{$\text{MHC}_\text{NC}$} that is trained without any curriculum on the full set of directives (locomotion, partial and whole-body). We trained \textit{$\text{MHC}_\text{NC}$} until the training loss plateaued for a long period of time, providing substantially more training time than the MHC. In Table \ref{table:curr_results} we compare \textit{$\text{MHC}_\text{NC}$}'s performance against MHC in perturbed standing, walking, and mimicry. We see that \textit{MHC} clearly outperforms \textit{$\text{MHC}_\text{NC}$} on the tasks of walking and mimicry, which shows the effectiveness of our curriculum design. It is interesting to note that \textit{$\text{MHC}_\text{NC}$} slightly outperforms the MHC with respect to standing against perturbation. This appears to come with the trade-off of not aligning itself to the command directive as well as the MHC, particularly in the lower body.

\begin{figure}[t]
  \begin{center}
    \includegraphics[width=\linewidth]{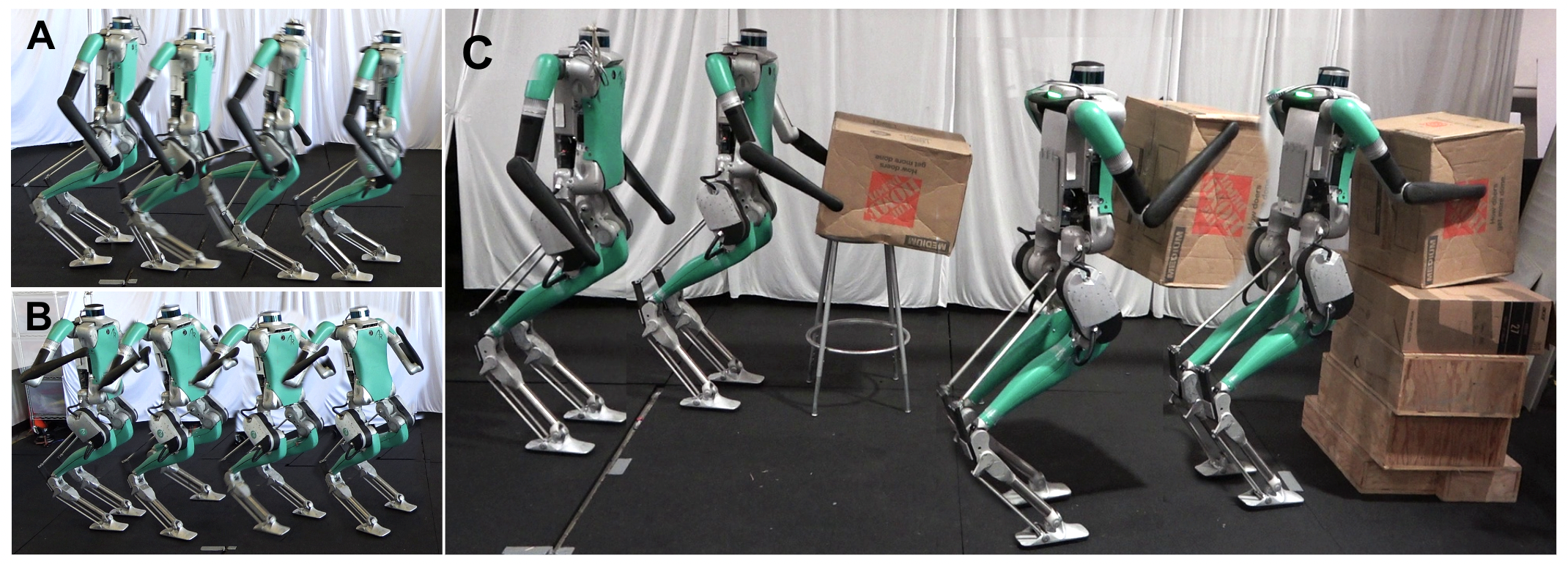}
  \end{center}
  \caption{
    Real-world demonstrations of our approach. A) Locomotion Directives using joystick commands. B) Fully-specified directive for a boxing jab; showing whole-body torso motion and foot coordination. C) Handcrafted sequence of masked directives combining upper body motion trajectory and lower body joystick commands; showing the ability to move to a location, pick up a box, move while holding it, and place it at a target location.} 
  \label{figure:sim2real}
\end{figure}

\renewcommand{\arraystretch}{1.3}
\begin{table}[t]
\setlength{\tabcolsep}{2mm} % Slightly compress columns if needed
\begin{tabular}{|r@{}l|l||l|l|}
\hline
         &              & \multicolumn{1}{c||}{\textbf{Metric}}& \multicolumn{1}{c|}{\textbf{MHC}} & \multicolumn{1}{c|}{\textbf{$\text{MHC}_\text{NC}$}} \\ \hline
\textbf{Stand} &\ \textbf{(50 N)}      & Root$_\Delta$(in m)           & 0.256                    & \textbf{0.189}                             \\
         &\ \textbf{(100 N)}     & Root$_\Delta$(in m)           & 0.311                    & \textbf{0.234 }                            \\ \hline
\textbf{Walk}  &\ \textbf{(1x m/s)}   & Root$_\Delta$(in m)           & \textbf{0.413}                   & 1.346                             \\
         &\ \textbf{(0.5y m/s)} & Root$_\Delta$(in m)           & \textbf{0.343}                 & 0.937                             \\ \hline
         &              & $\mathbb{E}_{\text{MEPE}}$(in m)& \textbf{0.098 }                   & 0.149                             \\
\textbf{Mimic}  &              & Root$_\Delta$(in m)           & \textbf{0.141 }                   & 0.327                             \\
         &              & Fail \%                       & \textbf{0.450}                    & 4.875                             \\ \hline
\end{tabular}
\caption{
    Performance metrics comparing a model trained with curriculum (MHC) against a model trained without curriculum ($\text{MHC}_\text{NC}$) on the tasks of standing, walking and mimicry (averaged for both partial and whole-body directives from motions in dataset \textit{setA}).
}
\label{table:curr_results}
\end{table}

\subsection{Sim-to-Real Results} 
We conducted real-world demonstrations of the MHC using the Digit V3 humanoid robot. Figure \ref{figure:sim2real} shows trials including box locomanipulation, which involved the input modality of locomotion and upper-body end-effector tracking. For a quantitative evaluation we tested the 20 motions in \textit{setA} on the real robot, performing 3 repetitions of each motion. The results are given in Table \ref{table:input_result}. While the failure rate (5\%) and joint position error are higher than observed in simulation, these results indicate our approach is able to achieve highly non-trivial sim-to-real transfer.

\section{Future Work}
This work motivates a number of important research directions: 1) Addressing the remaining sim-to-real gaps, especially for motions with wide foot placements or extended single-foot balancing stances. Incorporating real-world data into the sim-to-real process is likely to play an important role. 2) Continuing to increase the number and types of motions used for MHC training. Currently, for Digit V3, the bottleneck is our reliance on the MuJoCo simulator, compared to other higher throughput simulators such as Isaac Sim, which do not yet support the Digit V3 model due to modeling complexities. 3) Increasing the variety of masking patterns used for training to further enhance versatility. 4) Handling inconsistent and/or impossible input directives (e.g. root velocity is inconsistent with other joint components), which are currently not part of training.

\section{Acknowledgments}
This work is supported by NSF Award 2321851 and DARPA contract HR0011-24-9-0423.

\bibliography{aaai25}

\end{document}